\documentclass[11pt,a4paper]{article}
\usepackage[hyperref]{acl2021}
\usepackage{amsfonts}
\usepackage{amsmath, amssymb}
\usepackage{times}
\usepackage{latexsym}

\usepackage{microtype}

\aclfinalcopy 
\def\aclpaperid{694} 

\usepackage{url}            
\usepackage{booktabs}       
\usepackage{amsfonts}       
\usepackage{amsmath}
\usepackage{nicefrac}       
\usepackage{graphicx}
\usepackage{microtype}      
\usepackage{tabularx}
\usepackage{xcolor}
\usepackage{bbm}
\usepackage{array}
\usepackage{arydshln}
\usepackage{amsfonts}
\usepackage{amsmath}

\DeclareMathOperator*{\argsort}{arg\,sort}
\usepackage{bbm}
\usepackage{boldline}
\usepackage{bigstrut}
\usepackage{blindtext}
\usepackage{booktabs, siunitx}
\usepackage[labelfont=bf, format=plain, justification=justified, singlelinecheck=false]{caption}
\usepackage{color}
\usepackage{cprotect}
\usepackage{ctable}
\usepackage{dirtytalk}
\usepackage{enumitem}
\usepackage[export]{adjustbox}
\usepackage{float}
\usepackage{graphicx}
\usepackage{hhline}
\usepackage{latexsym}
\usepackage{mathrsfs}
\usepackage{microtype}
\usepackage{moresize}
\usepackage{multicol}
\usepackage{multirow}
\usepackage{nccmath}
\usepackage{nicefrac}
\usepackage{pifont}
\usepackage{placeins}
\usepackage{soul}
\usepackage{subcaption}
\usepackage{times}
\usepackage[utf8]{inputenc}
\usepackage{url}
\usepackage{verbatim}
\usepackage{wrapfig, lipsum}
\usepackage{textcomp}
\usepackage{enumitem}

\newcommand{\cut}[1]{}

\newcommand{\xmark}{\ding{55}}
\newcommand{\cmark}{\ding{51}}

\title{End-to-End Training of Neural Retrievers for \\ Open-Domain Question Answering}

\author{
Devendra Singh Sachan$^{1,2}$\Thanks{ This work was done during an internship at NVIDIA. Corresponding authors: Devendra Sachan, Mostofa Patwary.}, Mostofa Patwary$^{3}$, Mohammad Shoeybi$^{3}$, Neel Kant$^{3}$, \\
{\bf Wei Ping$^{3}$, William L Hamilton$^{1,2, 4}$, Bryan Catanzaro$^{3}$} \\
$^{1}$Mila - Quebec AI Institute; 
$^{2}$McGill University; 
$^{3}$NVIDIA;
$^{4}$Canada CIFAR AI Chair\\
{\tt sachande@mila.quebec, mpatwary@nvidia.com}\\
}

\date{}

\begin{document}
\maketitle


\begin{abstract}
Recent work on training neural retrievers for open-domain question answering (OpenQA) has employed both supervised and unsupervised approaches. 
However, it remains unclear how unsupervised and supervised methods can be used most effectively for neural retrievers.
In this work, we systematically study retriever pre-training. 
We first propose an approach of unsupervised pre-training with the Inverse Cloze Task and masked salient spans, followed by supervised finetuning using question-context pairs.
This approach leads to absolute gains of $2$+ points over the previous best result in the top-20 retrieval accuracy on Natural Questions and TriviaQA datasets.
We next explore two approaches for end-to-end training of the reader and retriever components in OpenQA models, which differ in the manner the reader ingests the retrieved documents. 
Our experiments demonstrate the effectiveness of these approaches as we obtain state-of-the-art results. On the Natural Questions dataset, we obtain a top-20 retrieval accuracy of 84\%, an improvement of $5$ points over the recent DPR model.
We also showcase good results on answer extraction, outperforming recent models such as REALM and RAG by $3$+ points. 
Our code is available at:  \url{https://github.com/NVIDIA/Megatron-LM}.
\end{abstract}


\section{Introduction}

\begin{figure}[t]
\centering
\includegraphics[max width=0.5\textwidth, scale=0.9]{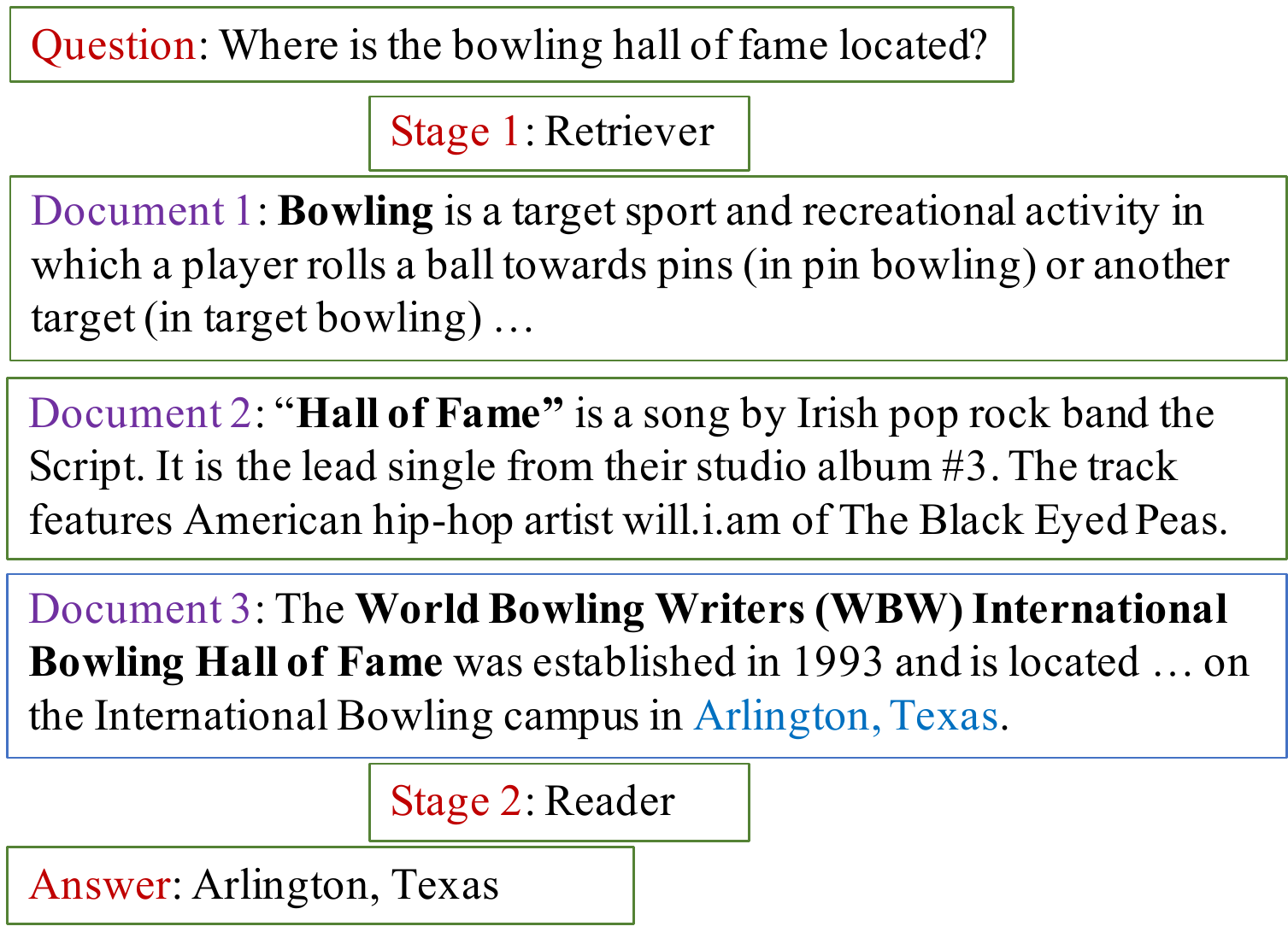}
\vspace{-8pt}
\caption{An example illustrating OpenQA pipeline.}
\label{fig:openqa-intro}
\vspace{-7pt}
\end{figure}

The task of open-domain question answering (OpenQA) consists of finding \emph{answers} to the information-seeking \emph{questions} using a large knowledge source such as Wikipedia. 
This knowledge source is also referred to as \emph{evidence} and it typically contains millions of documents.
Most approaches for OpenQA consist of a two-stage pipeline~\cite{chen2017reading,chen2018neural}. In the first stage, given a question, a \emph{retriever} module identifies the most relevant documents, which is often a very small subset of the evidence known as \emph{context documents}. Traditionally, approaches based on document ranking such as BM25~\cite{Robertson2009bm25} have been used for the retriever. In the second stage, these relevant documents are given as input to the \emph{reader} module, which understands them and extracts the answer for the question (Figure~\ref{fig:openqa-intro}).

The main drawback of the BM25 method is that it is not trainable and hence it can't be adapted to tasks involving open-retrieval.
Recent work has addressed this limitation by building upon advances in self-supervised learning, 
such as BERT~\cite{devlin2019bert}.
These approaches model both the retriever and reader using neural networks, allowing the retriever to be trained using task-specific datasets~\cite{lee-etal-2019-latent,guu2020realm}.
Typically, the retriever model consists of a \emph{dual-encoder} architecture~\cite{bromley1994signature}, where one encoder processes the question and the other encoder processes the context document. 
Prior work has investigated both unsupervised and supervised approaches to train the retriever.
Unsupervised approaches include separately training the retriever with Inverse Cloze Task (ICT)~\cite{lee-etal-2019-latent} or training the retriever and reader jointly by predicting masked salient spans (REALM)~\cite{guu2020realm}, while supervised approaches such as Dense Passage Retrieval (DPR)~\cite{karpukhin2020dense} train the retriever using human-annotated sets of question and context pairs. 

However, there is no study that investigates the comparative advantages of using these two styles of training when the retrieval task is challenging, \emph{i.e.}, when the evidence contains millions of documents. 
It is unclear if the unsupervised approaches can further help to improve the performance of \emph{strong} supervised approaches, and, if so,  under what conditions.
A core focus of this work is systematically studying these aspects of retriever training.

We propose a unified approach to train the retriever: unsupervised pre-training followed by supervised finetuning. 
We also investigate key design choices---such as relevance score scaling and longer training---and showcase their effectiveness. 
Our results demonstrate that the proposed approach obtains substantial accuracy gains when evaluated on benchmark OpenQA datasets. Extensive experiments also highlight the relative importance of different pre-training strategies, revealing important trade-offs when varying the amount of supervised data available to train the retriever.

Furthermore, motivated by recent work~\cite{guu2020realm,lewis2020pre}, we also explore two approaches for end-to-end supervised training of the reader and retriever components. In the first approach, the reader considers each retrieved document separately while in the second approach, the reader takes as input all the retrieved documents together. We compare the effectiveness of these approaches on both retrieval accuracy and answer extraction. We show that the first approach leads to an improved retrieval performance, while the second approach results in an improved answer extraction. With end-to-end training, we outperform previous best models to obtain new state-of-the-art results on retrieval accuracy and answer extraction. We also perform experiments by scaling the model size to a large configuration for both retriever and reader and observe consistent improvements, compared with smaller models.

In summary, the contributions of this work are:
\vspace{-2mm}
\begin{itemize}[noitemsep,leftmargin=*]
    \item We demonstrate that our proposed method of \emph{unsupervised pre-training} of the retriever with ICT followed by \emph{supervised finetuning} leads to absolute gains of more than 2 points in the top-20 retrieval accuracy over the previous best result on Natural Questions and TriviaQA datasets.
    \item We show that \emph{masked salient spans}-based pre-training of the retriever is more effective when the supervised dataset sizes are small.
    \item Our \emph{end-to-end} training approach obtains new state-of-the-art performance on retrieval accuracy. On Natural Questions, our top-20 accuracy is $84$, which is a $5$ points gain over DPR results. 
    \item We achieve competitive results on \emph{answer extraction} with gains of more than $3$ points over recent models such as REALM~\cite{guu2020realm} and RAG~\cite{Lewis2020Retrieval}.
    \item We \emph{scale up} end-to-end training to \emph{large models} and show \emph{consistent gains} in performance.
\end{itemize}

The rest of the paper is organized as follows. Sec.~\ref{sec:retriever} and \ref{sec:end-to-end-training} explain the retriever model and end-to-end training, respectively. Sec.~\ref{sec:exp-setup}-\ref{sec:answer-extraction} describe  the  experimental  details with the results. Sec.~\ref{sec:related-work} reviews the related work followed by conclusion in Sec.~\ref{sec:conclusion}.


\section{Neural Retriever} \label{sec:retriever}
In this section, we first describe the retriever architecture and then discuss different approaches to train it, including our proposed approach.

\subsection{Background}
Given a collection of documents in the evidence $\mathcal{Z}=\{z_1, \cdots, z_m\}$ and a question $q$, the task of the retriever is to select a relevant subset of documents for the question.
To do this, the retriever performs a ranking of the evidence documents conditioned on the question and outputs the top-ranked documents.

The retriever model consists of two modules: a question encoder ($f_Q$) and a context encoder ($f_Z$). Such a model is often referred to as a \emph{dual-encoder model}~\cite{bromley1994signature}. 
Here, we detail the training methodology of the dual-encoder model given a questions ($q$) and context documents ($z_i$) from $\mathcal{Z}$. First, we compute the \emph{relevance score} between the question and context. We define the relevance score to be the dot-product between the question and context representations
\begin{equation}
s(q, z_i; \phi) = f_Q(q)^\top f_Z(z_i)
\end{equation}
where $f_Q(q) \in \mathbb{R}^d$ and $f_Z(z) \in \mathbb{R}^d$ denote 
the question and context encoders, respectively, which are parameterized by $\phi = [\phi_Q, \phi_Z]$.
We model the $f_Q$ and $f_Z$ using BERT-style transformer networks~\cite{devlin2019bert,vaswani2017attention}. We consider the hidden states of the first token of the sequence (i.e.\ [CLS] token) as the encoder's output. 
The probability of a context document $z_i$ being relevant to the question $q$ is calculated as

\begin{equation} \label{eq:ret-sim}
p(z_i \mid q, \mathcal{Z}; \phi) = \frac{\exp(s(q, z_i; \phi)/\tau)} {\sum_{j=1}^{|\mathcal{Z}|} \exp(s(q, z_j; \phi)/\tau)}
\end{equation}
where $\tau$ is a scaling factor. While previous work had used the setting of $\tau=1$, in this work, we set $\tau=\sqrt{d}$. Bigger scaling factor helps in better optimization when the model hidden size ($d$) is large. We refer to this as \emph{relevance score scaling}. To train the retriever, we maximize the log-likelihood computed from Eq.~\ref{eq:ret-sim}.

In practice, as the evidence set consists of millions of documents, the normalization term would be expensive to compute. Hence, we approximate the denominator of the above equation by using the context documents in the batch as negative examples, a technique that has shown to perform well in practice~\cite{pmlrchen20j}.

\subsection{Training}
In this section, we discuss different approaches to train the retriever. In all the approaches, we initialize the parameters of both the question and context encoders using BERT weights as implemented in Megatron-LM~\cite{shoeybi2019megatron}. We also experimented with random initialization but it vastly underperformed BERT initialization.

\subsubsection{Supervised Training} \label{methods:supervised} 
In the supervised setting,  \emph{human-annotated} questions, answers, and sometimes context are provided. If the context is not included, then a common approach is to use distant supervision~\cite{mintz2009distant} to obtain the context document. Specifically, we select the top-ranked document using BM25~\cite{Robertson2009bm25} from the evidence that contains the answer as the context. We also select other top-ranked documents that do not contain the answer as additional \emph{hard} negative examples. This approach to train neural retriever was popularized by~\cite{karpukhin2020dense}.

\subsubsection{Unsupervised Training} \label{subsec:unsup-ret-train}

\paragraph{Inverse Cloze Task (ICT):} 
In this setup, we do not consider the human-annotated question-context pairs. 
Instead, the retriever is trained in an unsupervised manner. 
Specifically, a randomly sampled sentence from a paragraph is considered as the query while other sentences as the context. This approach was first proposed by~\cite{lee-etal-2019-latent}.

\paragraph{Masked salient spans training:} 
~\cite{guu2020realm} showcased that the ICT initialized retriever can be further improved by training it with an objective where the reader predicts the masked salient spans such as named entities conditioned on the retrieved documents. In this work, we adopt the same approach. However, unlike~\cite{guu2020realm} who use BERT for the reader, we use a generative language model based on T5~\cite{raffel2020t5}.

\subsection{Proposed Approach: Unsupervised Pre-training and Supervised Finetuning} \label{proposed-approach}
To improve the retriever training, we propose the approach of unsupervised pre-training of the retriever followed by supervised finetuning. In this approach, we first pre-train the retriever weights with ICT training or masked salient spans training (Sec.~\ref{subsec:unsup-ret-train}). 
After pre-training, we finetune the retriever with supervised training (Sec.~\ref{methods:supervised}).

\section{End-to-End Retriever and Reader Training} \label{sec:end-to-end-training}

\begin{figure*}[t]
\centering
\includegraphics[max width=1.0\textwidth, scale=0.9]{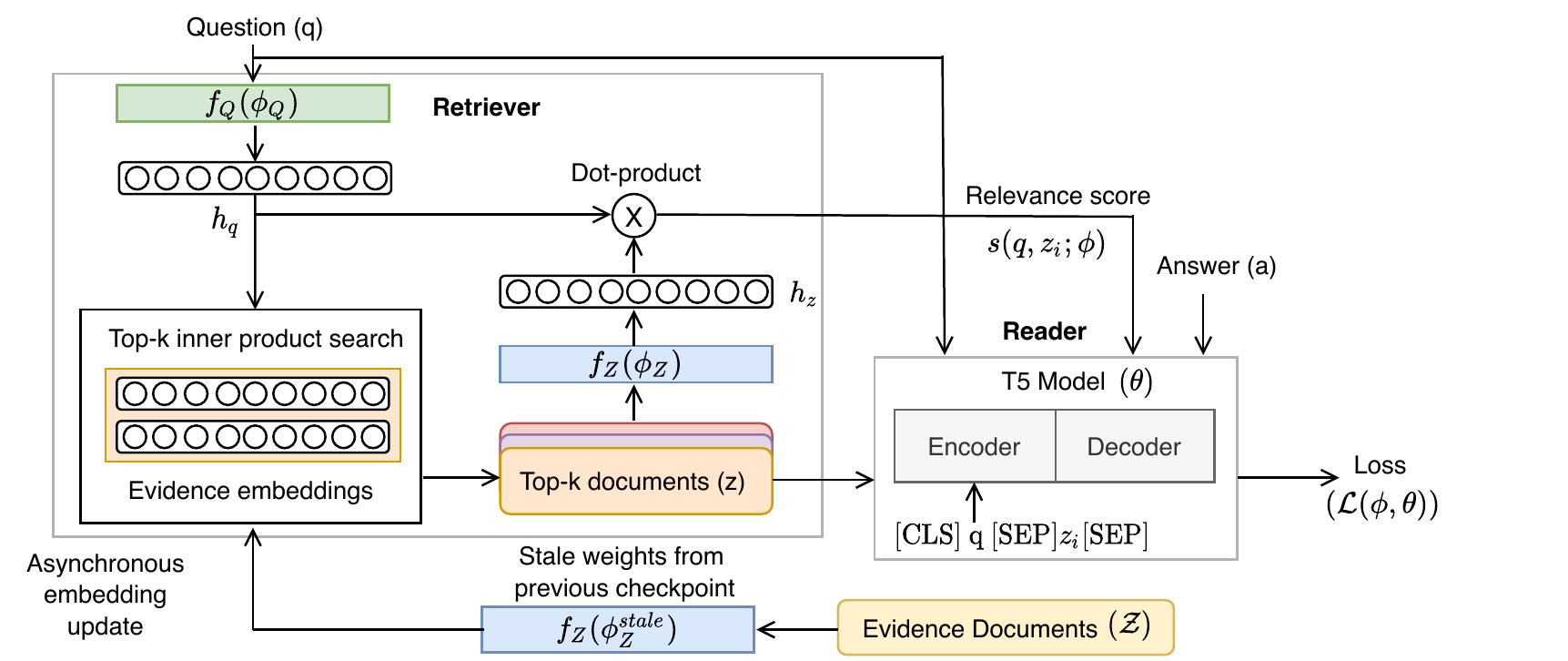} \hspace{-2cm}
\vspace{-5pt}
\caption{A schematic diagram illustrating end-to-end supervised training of the retriever and reader components.}
\label{fig:end-to-end-training}
\vspace{-6pt}
\end{figure*}

In this section, we  explore two \emph{supervised training} approaches to end-to-end train the reader and retriever components from the task-specific data. In the first approach, the reader considers each retrieved document separately (Sec.~\ref{sim-weighted_likelihood}) while in the second approach, the reader takes as input all retrieved documents together (Sec.~\ref{inter-attention}). These approaches are designed such that when predicting the answer conditioned on the question, the learning process improves both the reader and retriever.

\paragraph{Background and notation:} In end-to-end training, the trainable components consists of the retriever ($\phi$) and reader ($\theta$) parameters. For retriever, we use the dual-encoder architecture and train it as discussed previously in Sec.~\ref{proposed-approach}. 
Our reader is a \emph{generative model} designed according to the sequence-to-sequence modeling paradigm~\cite{ilya2014sequence}. Specifically, we use pre-trained T5 as the reader.
The inputs to the training process are \emph{questions} ($q$) and its \emph{answers} ($a$), both in string form.
Given a question, first the retriever obtains the \emph{k relevant} context documents ($\mathcal{K}$) from the evidence ($\mathcal{Z}$) as
\begin{equation}
\mathcal{K} = \argsort_{z_i \in \mathcal{Z}} s(q, z_i; \phi)[:k]
\end{equation}
The reader then takes the question and one or more context documents ($z_i$) as input to predict the answer, the likelihood of which is defined as
\begin{equation}
p(a \mid q, z_i; \theta) = \prod_{j=1}^{N} p\left(a_j \mid a_{1:j-1}, q, z_i; \theta\right),
\end{equation}
where $N$ is the number of answer tokens. Next, we describe the two proposed approaches. A block diagram illustrating the end-to-end training process is shown in Figure~\ref{fig:end-to-end-training}.

\subsection{Approach 1: Individual Top-k} \label{sim-weighted_likelihood}
In this approach, similar to~\cite{guu2020realm}, the reader's likelihood is first computed conditioned on the question and each retrieved document. The marginal likelihood is defined as the weighted average of the \emph{individual} likelihoods as

\begin{equation} \label{eq:indiv-topk}
p(a \mid q; \theta,\phi) = \sum_{z_i \in \mathcal{K}} p(a \mid q, z_i; \theta) p(z_i \mid q, \mathcal{Z}; \phi),
\end{equation}
where $p(z_i \mid q, \mathcal{Z}; \phi)$ is computed using Eq.~\ref{eq:ret-sim}. However, the normalization is done over $\mathcal{K}$ instead of $\mathcal{Z}$. The final loss is defined as the negative marginal log-likelihood
\begin{equation} \label{eq:loss}
\mathcal{L}(q, a) = -\log p(a \mid q; \theta,\phi).
\end{equation}
We note that the RAG model~\cite{Lewis2020Retrieval} also proposed a similar approach, but there are two main differences. The first is that while we update all the parameters of the retriever (both the query and context encoders), RAG just updates the query encoder. The second is that we use T5 model as the reader while RAG uses BART model~\cite{lewis2020bart}. These enhancements help us obtain substantial gains over the RAG model, which we will discuss in Sec.~\ref{sec:answer-extraction}.

\subsection{Approach 2: Joint Top-k} \label{inter-attention}

In this approach, similar to~\cite{lewis2020pre}, the likelihood is defined as the reader's likelihood conditioned on the question, \emph{all} the retrieved documents, and the retrieval score

\begin{equation} \label{eq:joint-topk}
p(a \mid q; \theta,\phi) = p(a \mid q, z_{1:k},p(z \mid q, \mathcal{Z}; \phi); \theta).
\end{equation}

As the T5 reader consists of separate encoder and decoder modules, it provides the flexibility to customize the input or output of the encoder. 
We concatenate each retrieved document with the question and feed them as input to the encoder, which computes their hidden representations. 
Next, we stack the hidden representations of all the retrieved documents, which the decoder \emph{jointly} attends to during the encoder-decoder attention, thus allowing a more powerful form of information aggregation from multiple retrieved documents. 
We also add retriever similarity score to bias the encoder-decoder attention as it helps facilitate end-to-end training and enables the reader to pay higher attention to the relevant documents.
The interaction score during the encoder-decoder attention is computed as

\begin{equation}
\text{attn}(q, a, z_{1:k}) \propto Q(a)^\top K(z_{1:k}, q) + \lambda p(z \mid q; \phi),
\end{equation}
where $Q$ is the query vector computed from decoder's input, $K$ is the key vector computed from encoder's output, and $\lambda$ is a trainable parameter. 

Final loss is defined according to Eq.~\ref{eq:loss}. We further note that a similar approach for OpenQA was proposed in~\cite{izacard2020leveraging} but it only optimizes the reader model and didn't perform end-to-end training of the retriever.


\section{Experimental Setup} \label{sec:exp-setup}
In this section, we describe the datasets and model settings. For reproducibility, we provide training details and list the hyperparameters in Appendix~\ref{sec:training_details}.

\subsection{OpenQA Datasets}
We perform experiments using two widely used QA datasets whose details are provided below and their statistics are shown in Table~\ref{tab:dataset_stats}.

\paragraph{Natural Questions (NQ):} This corpus consists of real questions asked from the Google search engine along with their long and short answer annotations from the top-ranked Wikipedia pages~\cite{Kwiatkowski2019natural}. Following prior work~\cite{karpukhin2020dense}, we use the same 
subset of the short answer questions in our experiments, as it is more suited for OpenQA.

\paragraph{TriviaQA:}
This corpus consists of a collection of trivia questions and their answers scraped from multiple sources in the Web~\cite{joshi2017triviaqa}. 

\paragraph{Evidence:}
Following~\cite{karpukhin2020dense}, we make use of their released preprocessed English Wikipedia dump from December 2018 as the source of evidence documents. Overall, there are $21,015,324$ documents, each $100$ words long.

\begin{table}[t]
\centering
\small
\begin{tabular}{@{}l c c c c@{}}
 \toprule
 \textbf{Dataset} & \textbf{\ Train} & \textbf{Filtered Train}  & \textbf{Dev} & \textbf{Test} \\
 \midrule
 NQ & 79,168 & 58,880 & 8,757 & \phantom{0}3,610 \\
 TriviaQA          & 78,785 & 60,413 & 8,837 & 11,313 \\
 \bottomrule
 \end{tabular}
 \vspace{-6pt}
\caption{OpenQA dataset statistics. The training set is used for end-to-end training, while the filtered version is used for retriever training. The filtered set ignores those examples where the document retrieved from evidence does not align with the ground-truth document.
}
\label{tab:dataset_stats}
\vspace{-6pt}
\end{table}

\subsection{Model Details} \label{sub:model_details}
We use two models of different sizes, {\it base} and {\it large}, for the experiments.
The base configuration consists of $12$ layers, $768$-d hidden size, and $12$ attention heads. The BERT-base contains $110$M parameters while the T5-base contains $220$M parameters.
The large configuration consists of $24$ layers, $1024$-d hidden size, and $16$ attention heads. The BERT-large contains $330$M parameters while the T5-large contains $770$M parameters.


\section{Results: Retriever Training} \label{results-ret-training}

\begin{table}[t]
\centering
\small
\begin{tabular}{@{}l c c c c@{}}
 \toprule
 \textbf{Setting} & \textbf{Top-1}  & \textbf{Top-5} & \textbf{Top-20} & \textbf{Top-100} \\
 \midrule
 \multicolumn{5}{c}{\textit{Base Configuration}}\\
 \midrule
 \lbrack CLS\rbrack, 40 epochs      & 32.6 &	60.1 & 76.4 & 85.9 \\
 + score scaling  & 34.1 & 60.9 &	77.6 & 85.9 \\
 + 80 epochs      & 36.7 & 62.2 & 77.4 & {\bf 86.0} \\
 + 1 hard negative  & {\bf 48.6} & {\bf 74.5} & {\bf 79.0} & 85.8 \\
 \midrule
 DPR (Official) & -- & 67.1 & 78.4 & 85.4 \\
 \bottomrule
 \end{tabular}
 \vspace{-6pt}
\caption{Effect of different factors on the supervised training of retriever when evaluated on NQ test set.}
\label{tab:score-scaling-gnq}
\vspace{-6pt}
\end{table}

\begin{table*}[t]
\small
\centering
\begin{tabular}{@{}l c c c c c c c c@{}}
 \toprule
 \textbf{Model} & \multicolumn{4}{c}{\textbf{NQ}} & \multicolumn{4}{c}{\textbf{TriviaQA}} \\
 \midrule
  & \textbf{Top-1}  & \textbf{Top-5} & \textbf{Top-20} & \textbf{Top-100} & \textbf{Top-1} & \textbf{Top-5} & \textbf{Top-20} & \textbf{Top-100} \\
 \midrule
 \multicolumn{9}{c}{\textit{Base Configuration}}\\
 \midrule
 BERT (zero-shot)            & \phantom{0}0.9  & \phantom{0}3.9 & \phantom{0}9.4 & 20.3 & \phantom{0}0.6 & \phantom{0}2.8 & \phantom{0}7.2 & 17.8 \\
 ICT (zero-shot)             & 12.6 & 32.3 & 50.6 & 66.8 & 19.2 & 40.2 & 57.5 & 73.6 \\
 Masked salient spans (zero-shot)     & 20.0 & 41.7 & 59.8 & 74.9 & 31.7 & 53.3 & 68.2 & 79.4 \\
 BERT + Supervised                & 48.6 & 68.8 & 79.0 & 85.8 & 57.5 & 72.2 & 80.0 & 85.1 \\
 ICT + Supervised          & 48.4 & {\bf 72.1} &  81.8 & {\bf 88.0} & 58.4       & 73.9       &  81.7      & 86.3 \\
 Masked salient spans + Supervised & {\bf 50.3} & 71.9 & {\bf 82.1}       & 87.8       & {\bf 60.6} & {\bf 74.8} & {\bf 81.8} & {\bf 86.6} \\
 \midrule
 \multicolumn{9}{c}{\textit{Large Configuration}}\\
 \midrule
 ICT (zero-shot)    & 13.0 & 31.8 & 49.3 & 66.1 & 20.1 & 41.6 & 58.5 & 74.1 \\
 BERT + Supervised    & 51.4 & 71.0 & 81.0 & 87.2 & 60.4 & 74.5 & 81.4 & 86.0 \\
 ICT + Supervised     & {\bf 52.4} & {\bf 72.7} & {\bf 82.6} & {\bf 88.3} & {\bf 61.9} & {\bf 76.2} & {\bf 82.9} & {\bf 87.1} \\
 \bottomrule
 \end{tabular}
 \vspace{-6pt}
\caption{Effect of unsupervised pre-training on retrieval accuracy when evaluated on NQ and TriviaQA test sets.
}
\label{tab:ret-init}
\vspace{-6pt}
\end{table*}

In this section, we compare different approaches to train the retriever. Retrieval accuracy is evaluated using the top-k metric ($k \in \{1, 5, 20, 100\}$).

\subsection{Effect of Relevance Score Scaling, Longer Training, and Hard Negatives}

We explore the best training settings for \emph{supervised} training of the retriever. 
To do so, we perform a series of experiments on the NQ dataset starting with the training settings from the popular DPR model and then progressively improve it. 
DPR was initialized with BERT, trained for $40$ epochs, with a scaling factor of $1$, and utilized [CLS] token embeddings from the retriever. 
Our result with this setting is shown in Table~\ref{tab:score-scaling-gnq}.
We then observe that incorporating relevance score scaling and longer training till $80$ epochs helps to improve the top-5 and top-20 accuracy by $1.5$-$2$ points.
These results also signify that the original DPR model was significantly undertrained and not fully optimized.

In addition to score scaling, we further include $1$ additional hard-negative example (similar to DPR) for each question-context pair and train the model for 80 epochs.
Our results, in sync with the results of DPR, obtain substantial additional gains in performance.
\emph{These findings highlight that relevance score scaling, longer training, and including a hard negative example are essential to improve the supervised retriever's accuracy}.
These supervised training results can be considered as a very strong baseline.
Hence, we employ these settings in subsequent experiments.

\subsection{Effect of Retriever Initialization} \label{subsec:ret-init}

We first characterize the zero-shot retriever's performance when its weights are initialized with either BERT or ICT or masked salient spans pre-training (Table~\ref{tab:ret-init}).
As is understood that unsupervised language models do not perform well in information retrieval tasks~\cite{lee-etal-2019-latent}, evidently, BERT also leads to a poor retrieval accuracy.
We note that ICT initialization is quite effective in providing a non-trivial zero-shot accuracy which is further improved by masked salient spans training by more than $8$ points. 
Both being unsupervised approaches demonstrate their utility in effectively bootstrapping the retriever almost from scratch.

We next empirically analyze our proposed approach of pre-training with ICT and masked salient spans followed by supervised finetuning.
We observe that it provides absolute improvements of $2$-$3$ points over the already strong supervised training results, with the gains being consistent across both the datasets.
These results highlight that even after finetuning the retriever with thousands of labeled examples, it does not lead to catastrophic forgetting of the discriminative properties learned by the retriever during ICT and masked salient spans pre-training.
Another merit is that being unsupervised, large text collections can be leveraged to pre-train the retriever, a considerable advantage over data-augmentation methods which rely on the availability of human-annotated question-context pairs. \emph{Furthermore, when comparing ICT with masked salient spans initialization, we note that their accuracy gains are roughly similar}.

\subsection{Effect of Amount of Training Data}

\begin{figure}[t]
\centering
\includegraphics[max width=0.5\textwidth, scale=0.9]{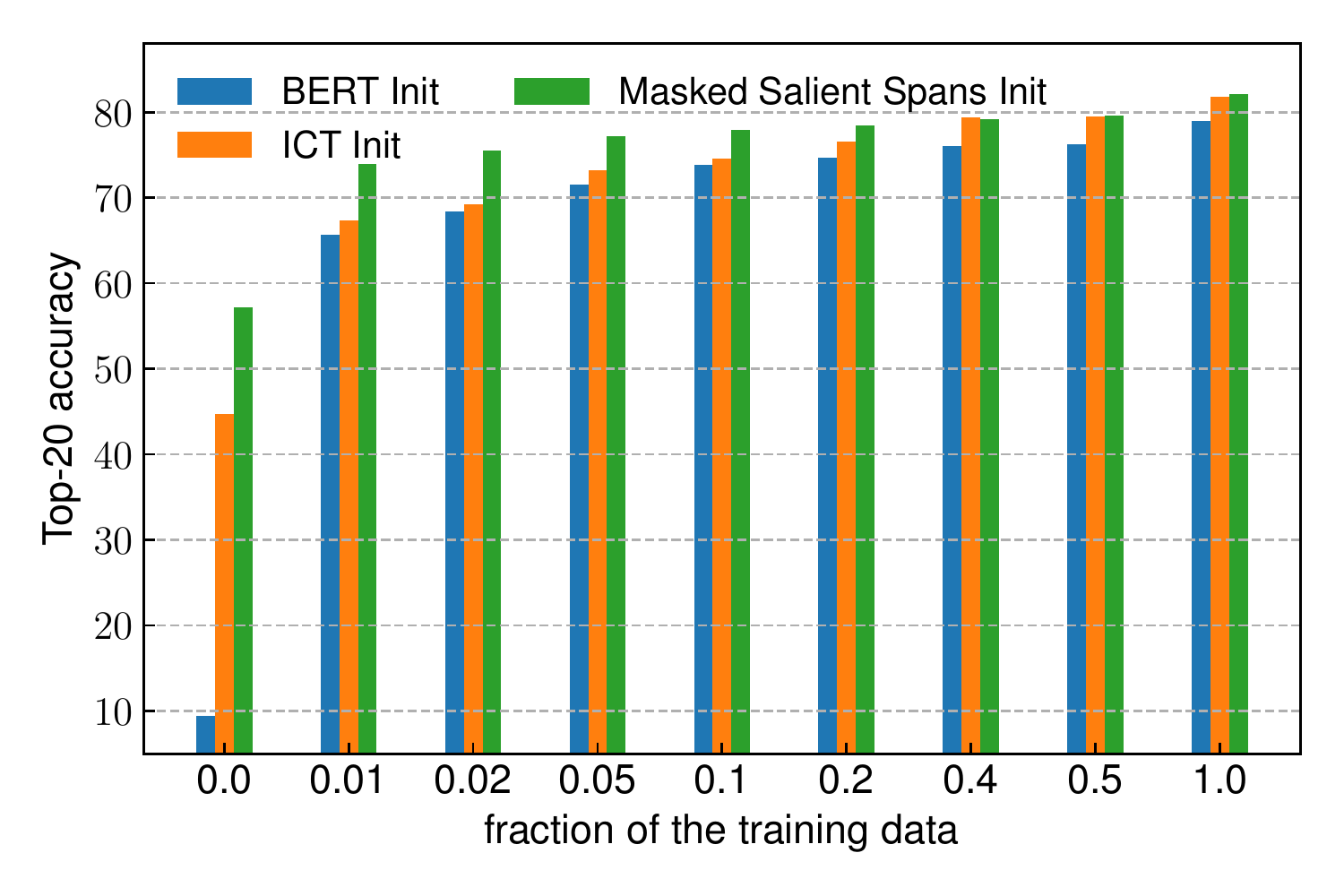}
\vspace{-10pt}
\caption{Effect of amount of training data on retrieval accuracy when evaluated on NQ test set.}
\label{fig:vary-train-nq}
\vspace{-6pt}
\end{figure}

We study the effect on accuracy when the retriever is pre-trained with BERT, ICT, or masked salient spans and the amount of supervised training data is varied. We train the retriever with $1\%$, $2\%$, $5\%$, $10$-$50\%$, of NQ's training data and plot the top-20 accuracy in Figure~\ref{fig:vary-train-nq}. 
Results reveal that in the low-resource regime, \emph{masked salient spans pre-training is much more effective than ICT, consistently leading to large gains}.
As the fraction of training data increases to beyond $40\%$ towards a high-resource setup, \emph{the gains from salient spans pre-training saturates to that of ICT}.
We believe that these findings will have important implications for future research in OpenQA---with only a few hundred examples, performing expensive masked salient span training is beneficial while if the training data has thousands of examples, ICT is just as optimal as masked salient spans training.

\subsection{Effect of End-to-End Training} \label{subsec:joint-ret}

\begin{table*}[t]
\small
\centering
\begin{tabular}{@{}l c c c c c c c c c c@{}}
 \toprule
 \textbf{Model} &  &  & \multicolumn{4}{c}{\textbf{NQ}} & \multicolumn{4}{c}{\textbf{TriviaQA}} \\
 \midrule
 & {\bf Q} & {\bf C} & \textbf{Top-1}  & \textbf{Top-5} & \textbf{Top-20} & \textbf{Top-100} & \textbf{Top-1}  & \textbf{Top-5} & \textbf{Top-20} & \textbf{Top-100} \\
 \midrule
 \multicolumn{11}{c}{\textit{Base Configuration}}\\
 \midrule
 DPR~\cite{karpukhin2020dense} & & & --   & 67.1 & 78.4 & 85.4 & --   &  --  & 79.4 & 85.0 \\
 \midrule
 ICT + Supervised     &  &  & 48.4 & 72.1 & 81.8 & 88.0 & 58.4 & 73.9 & 81.7 & 86.3 \\
 Individual Top-k     & \cmark & \xmark & 54.5 & 73.7 & 83.2 & 88.6 &  61.4    & 75.6 & 82.1 & 86.7 \\
 Individual Top-k     & \cmark & \cmark & {\bf 56.8} & {\bf 75.0} & {\bf 84.0} & {\bf 89.2} & {\bf 63.5} & {\bf 76.8} & {\bf 83.1} & {\bf 87.0} \\
 Joint Top-k     & \cmark & \xmark & 51.1 & 72.1 & 81.8 & 87.8 &  59.1 & 74.1 & 81.3 & 86.3 \\
 \midrule
 \multicolumn{11}{c}{\textit{Large Configuration}}\\
 \midrule
 ICT + Supervised     & & & 52.4 & 72.7 & 82.6 & 88.3 & 61.9 & 76.2 & 82.9 & 87.1 \\
 Individual Top-k & \cmark & \cmark & {\bf 57.5} & {\bf 76.2} & {\bf 84.8} & {\bf 89.8} & {\bf 66.4} & {\bf 78.7} & {\bf 84.1} & {\bf 87.8} \\
 Joint Top-k        & \cmark & \xmark & 53.7 & 73.3 & 83.2 & 88.0 & 61.2 & 75.9 & 82.7 & 87.0 \\
 \bottomrule
 \end{tabular}
 \vspace{-6pt}
\caption{Effect of end-to-end training using question-answer pairs on retrieval accuracy. {\bf Q} and {\bf C} signify if the query encoder and the context encoder are updated during training or not, respectively.}
\label{tab:joint-train-ret}
\vspace{-6pt}
\end{table*}

For end-to-end training, retriever weights are initialized with the previous best setting of ICT pre-training and supervised finetuning.
The number of retrieved evidence documents for the reader is considered as a hyperparameter and is selected via performance on the dev set.
The focus here is to analyze the effect on retrieval accuracy when updating the retriever weights using question-answer pairs in an end-to-end setting (Sec.~\ref{sec:end-to-end-training}). 
From the results in Table~\ref{tab:joint-train-ret}, we observe that for \emph{Individual Top-k}, when only the query encoder is updated, it tends to improve retrieval accuracy.
In addition, when the context encoder is also updated, the retrieval accuracy improves to $75\%$ at top-5, a big gain of $8$ points over the previous best DPR retriever.
Larger models further help to improve the performance leading to new state-of-the-art results. 

On the other hand, in \emph{Joint Top-k}, updating the query encoder just improves the top-1 score but does not really lead to much accuracy gains for higher top-k's.
We also do not update the context encoder for \emph{Joint Top-k} as it did not result in improvements during our initial experiments.

These results showcase that when the retriever is already well-initialized, the objective function of \emph{Individual Top-k} method is designed such that it significantly improves the retrieval accuracy while the \emph{Joint Top-k} method does not result in improvements. 
As we will show next, that the usefulness of this method lies in answer extraction.

\subsection{Intuition for Retriever Score Scaling}

\begin{table}[t]
\centering
\small
\begin{tabular}{@{}l c c c c c@{}}
 \toprule
 \textbf{$\times \sqrt{d}$} & \textbf{Top-1}  & \textbf{Top-5} & \textbf{Top-20} & \textbf{Top-100} & \textbf{Avg.} \\
 \midrule
 \multicolumn{6}{c}{\textit{Base Configuration}}\\
 \midrule
 0.25 & 48.8 & 69.3 & 78.7 & 85.5 & 70.6 \\
 0.5  & 51.4 & 71.6 & 81.5 & 87.7 & 73.1 \\
 1    & 51.1 & 71.8 & 82.1 & 87.7 & \textbf{73.2} \\
 2    & 50.2 & 71.5 & 81.9 & 87.9 & 72.9 \\
 4    & 50.6 & 71.7 & 81.7 & 88.0 & 73.0 \\
 \bottomrule
 \end{tabular}
 \vspace{-6pt}
\caption{Effect of score scaling factor ($\tau$) on the retrieval accuracy when evaluated on the NQ test set. The first column denotes the multiple ($m$) that is multiplied by $\sqrt{d}$ to obtain $\tau$, i.e., $\tau = m \times \sqrt{d}$ in Equation \ref{eq:ret-sim}.}
\label{tab:score-scaling-intuition}
\vspace{-6pt}
\end{table}

Retrieval score scaling is used when computing the probability distribution of the retrieved documents according to Equation \ref{eq:ret-sim}, where the retrieval score is normalized by the scaling factor ($\tau$). To study the effect of $\tau$ on the retrieval accuracy, we perform an ablation study with different values of $\tau$ on the NQ retrieval task, whose results can be seen in Table~\ref{tab:score-scaling-intuition}. More specifically, we choose different values of $\tau$ as a multiple of $\sqrt{d}$, where $d$ is the hidden size of the model. 
Our results indicate that the choice of $\tau=\sqrt{d}$ works well in practice.

Here, we briefly explain the intuition regarding the usage of the scaling factor. In our preliminary experiments on retriever training and end-to-end training without the scaling factor, we observed that a few of the top-k document’s similarity score with the query was very high that in turn led to it being assigned a high retrieval probability score. This high score was leading to a skewed probability distribution with most of the mass being centered over the top-1 or top-2 retrieved documents. A larger value of scaling factor results in a more even distribution of probability mass over the top-k documents, which in turn leads to better results in both retrieval accuracy and in the end-to-end training.

\section{Results: Answer Extraction} \label{sec:answer-extraction}
We next present the results of end-to-end training on answer extraction. 
To train the model, retriever weights are initialized with ICT pre-training and supervised finetuning while the reader is initialized with pre-trained T5 weights.
The number of retrieved evidence documents for the reader is tuned on the dev set.
Results are reported using the conventional Exact Match (EM) metric. 

\subsection{Individual Top-k Approach}

\begin{table}[t]
\small
\centering
\begin{tabular}{l c c}
 \toprule
 \textbf{Model} & \textbf{NQ} & \textbf{TriviaQA} \\
 \midrule
 \multicolumn{3}{c}{\textit{Base Configuration}} \\
 \midrule
 ORQA~\cite{lee-etal-2019-latent} & 33.3 & 45.0 \\
 REALM~\cite{guu2020realm} & 40.4 & -- \\
 DPR~\cite{karpukhin2020dense}   & 41.5 & 56.8 \\
 Individual Top-k & {\bf 45.9} & 56.3 \\
 \midrule
 \multicolumn{3}{c}{\textit{Large Configuration}} \\
 \midrule
 RAG~\cite{Lewis2020Retrieval} & 44.5 & 56.8 \\
 Individual Top-k & {\bf 48.1} & {\bf 59.6} \\
 \bottomrule
\end{tabular}
\vspace{-6pt}
\caption{Answer extraction results using \emph{Individual Top-k} approach. The grouping under base and large configurations is based on the size of the reader model.}
\label{tab:indiv-topk-ans}
\vspace{-6pt}
\end{table}

\begin{figure}[t]
\centering
\includegraphics[max width=0.5\textwidth, scale=0.95]{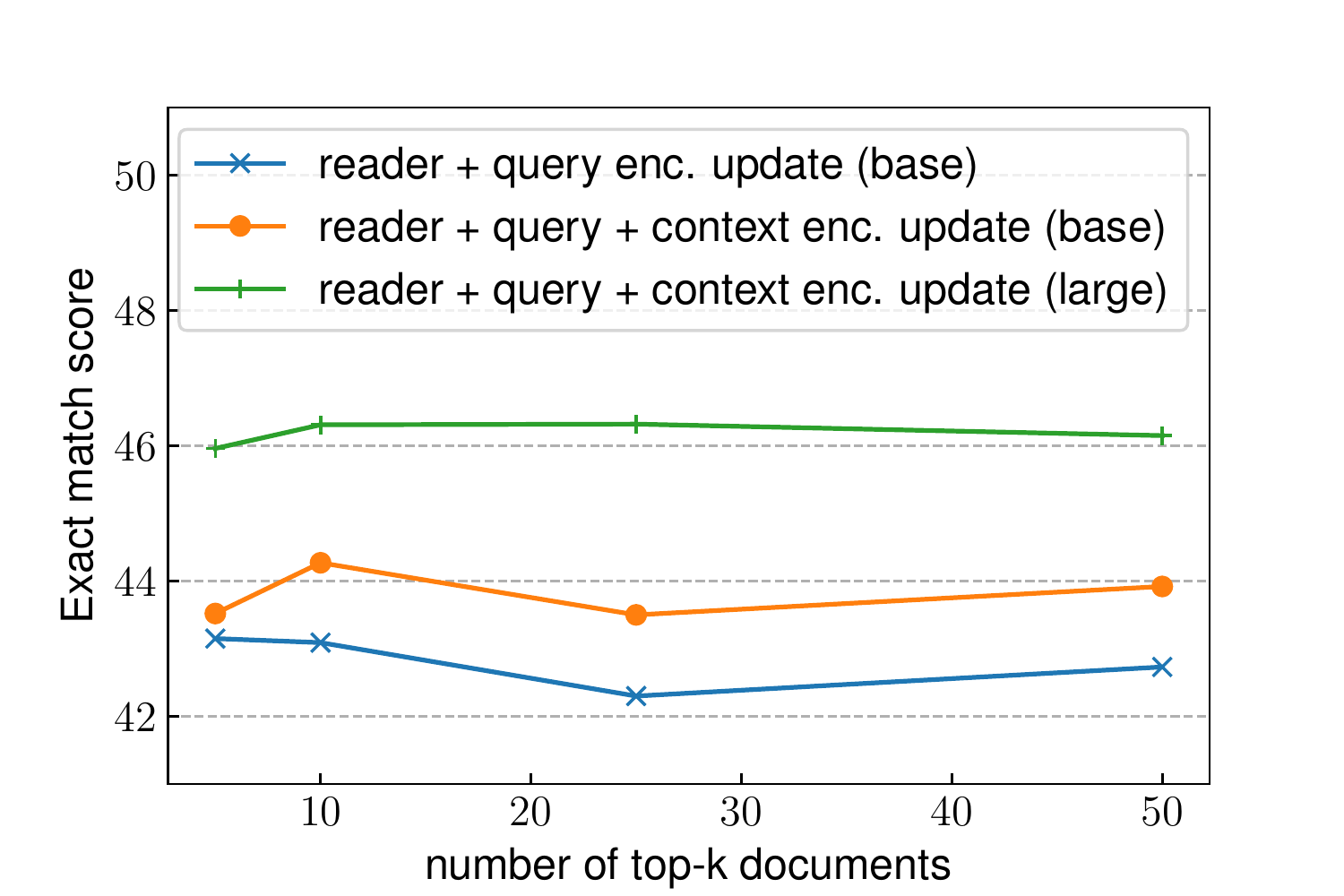}
\vspace{-8pt}
\caption{Effect of increasing top-k documents on answer generation for \emph{Individual Top-k} approach.}
\vspace{-6pt}
\label{fig:indiv-topk-varyk-nq}
\end{figure}

We compare our results as presented in Table~\ref{tab:indiv-topk-ans} with the recent related approaches in OpenQA.
For the base configuration on NQ, our model outperforms both REALM and DPR by more than $4$ points.
For the large configuration, we compare with the RAG model~\cite{Lewis2020Retrieval}, where
our approach outperforms it by $3.5$+ points on NQ and by $2.8$ points on TriviaQA.
\emph{Our improved results are because of a more accurate initial retriever, stronger reader, and updating both the query and context encoders during training}. 

Our analysis in Figure~\ref{fig:indiv-topk-varyk-nq} reveals that updating the context encoder improves the results for both the base and large configurations. Quite surprisingly, we also observe that the performance of \emph{Individual Top-k} approach is sensitive to the number of top-k documents and can also decrease with an increase in top-k documents. We leave an in-depth investigation of this as a future work.

\subsection{Joint Top-k Approach}

We compare our results with the recent Fusion-in-Decoder (FiD) approach~\cite{izacard2020leveraging} that also performs joint encoder-decoder attention. It consists of DPR as the retriever and T5 as the reader, which are initialized with their open-source weights.
However, unlike our approach, FiD just finetunes the reader weights.
Our results in Table~\ref{tab:joint-topk-ans} show that for the base configuration, Joint Top-k outperforms the FiD model by 1 point on NQ, highlighting the significance of end-to-end training.
For the large configuration, we obtain a gain of 0.7 points on TriviaQA. 

\begin{table}[t]
\small
\centering
\begin{tabular}{l c c}
 \toprule
 \textbf{Model} & \textbf{NQ} & \textbf{TriviaQA} \\
 \midrule
 \multicolumn{3}{c}{\textit{Base Configuration}} \\
 \midrule
 FiD~\cite{izacard2020leveraging} & 48.2 & 65.0 \\
 Joint Top-k & {\bf 49.2} & 64.8 \\
 \midrule
 \multicolumn{3}{c}{\textit{Large Configuration}} \\
 \midrule
 FiD~\cite{izacard2020leveraging} & 51.4 & 67.6 \\
 Joint Top-k & {\bf 51.4} & {\bf 68.3} \\
 \bottomrule
\end{tabular}
\vspace{-6pt}
\caption{Results on answer extraction using \emph{Joint Top-k} approach.
}
\label{tab:joint-topk-ans}
\vspace{-6pt}
\end{table}

\begin{figure}[!t]
\centering
\includegraphics[max width=0.5\textwidth, scale=0.95]{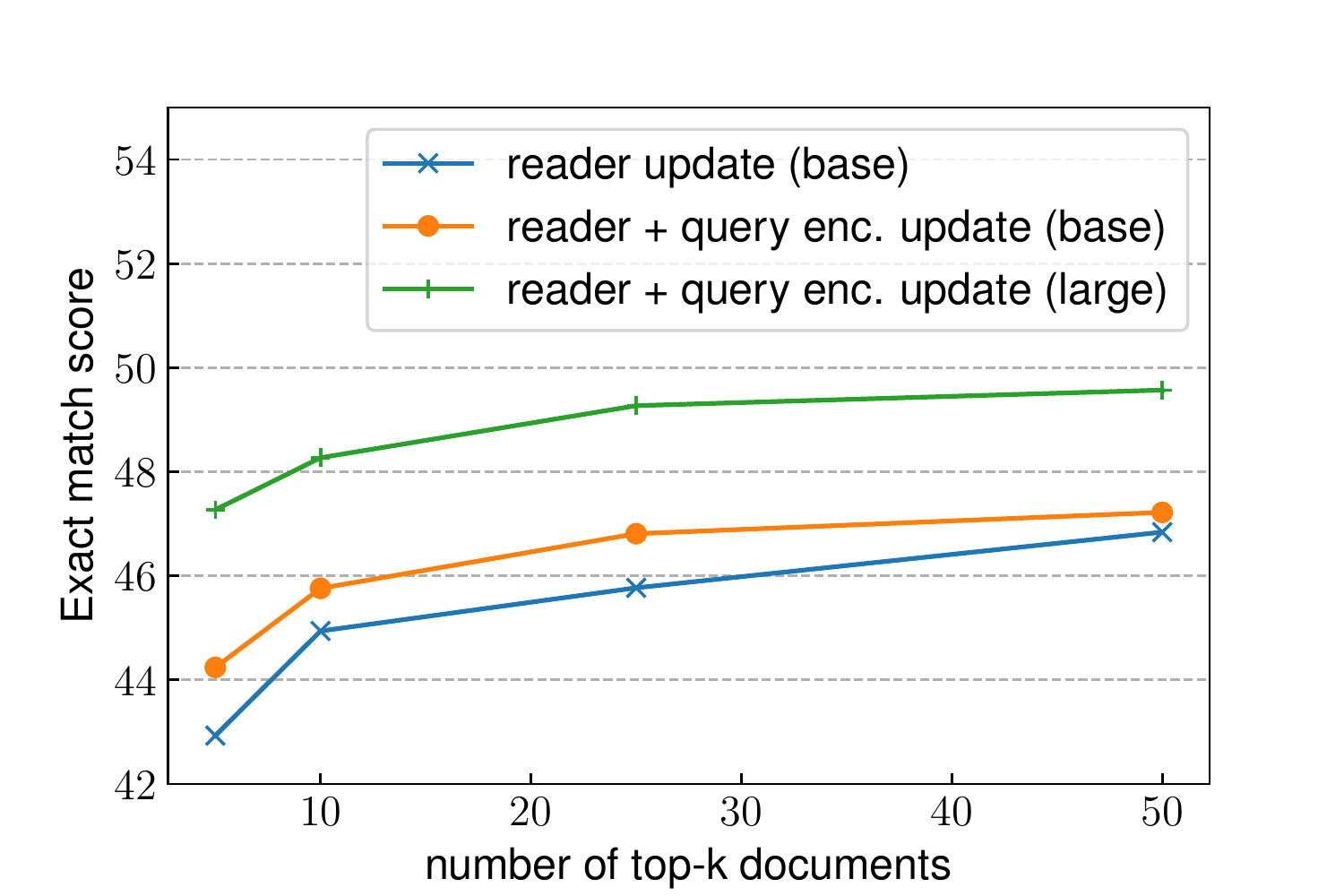}
\vspace{-8pt}
\caption{Effect of increasing top-k documents on answer generation for \emph{Joint Top-k} approach.}
\label{fig:joint-topk-varyk-nq}
\vspace{-6pt}
\end{figure}

Our analysis in Figure~\ref{fig:joint-topk-varyk-nq} portrays that the EM scores improve with more retrieved documents. This highlights that in contrast to \emph{Individual Top-k}, the \emph{Joint Top-k} better aggregates the information contained in the retrieved documents.
This Figure also illustrates the effect of similarity enriched attention on answer extraction for the base configuration. For values of top-k=5, 10, and 25, using retrieval-similarity enriched encoder-decoder attention, we consistently observe a gain of 0.8-1 EM points (comparing orange plot and blue plot in Figure~\ref{fig:joint-topk-varyk-nq}), while there is a smaller gain when top-k=50.
This signifies that with more retrieved documents, the utility of end-to-end training tends to diminish, thus explaining the lower gains observed in retrieval performance for \emph{Joint Top-k} in Table~\ref{tab:joint-train-ret}.

\subsection{Overall Comparison}
Based on the discussions in Sec.~\ref{subsec:joint-ret} and Sec.~\ref{sec:answer-extraction}, we remark that end-to-end training using the two approaches has a complementary effect on the retrieval accuracy and answer extraction. While the \emph{Individual Top-k} approach helps to significantly improve the retrieval performance, the \emph{Joint Top-k} approach is more useful for answer extraction. 

\section{Related Work} \label{sec:related-work}

\cite{yih-etal-2011-learning} proposed a discriminative approach to train a retriever by learning dense representations of query and context documents based on word frequency. However, this approach was data-hungry and not scalable.
Recently,~\cite{lee-etal-2019-latent,karpukhin2020dense} address this by leveraging pre-trained BERT weights~\cite{devlin2019bert} to train a dual-encoder retriever by using smaller amounts of question-context pairs. 
In particular,~\cite{lee-etal-2019-latent} first pre-train the retriever in an unsupervised manner using ICT and then jointly train the retriever and reader for OpenQA.
On the other hand,~\cite{karpukhin2020dense} perform supervised training of the retriever using hard-negative examples, yielding impressive results on several retrieval benchmarks.

To improve the retrieval accuracy of the dual-encoder model,~\cite{Chang2020Pre-training} explore several paragraph-level pre-training strategies including the application of ICT. They demonstrated the effectiveness of pre-training over sparse-retrieval approaches such as BM25. Their evidence consisted of the training documents that was further increased to $1$M documents for OpenQA. Our work differs from them in several ways. First, our OpenQA setup is more challenging as the evidence consists of $21$M documents. Second, we pre-train with two strategies consisting of ICT and masked salient-spans and finetune using strong supervised methods, which leads to much improved results. Third, we further update the retriever with end-to-end training leveraging question-answer pairs, which further improves the retrieval accuracy leading to new state-of-the-art results.

A new line of work investigates task-specific pre-training of language models. For example,~\cite{guu2020realm} predicts masked salient spans consisting of named entities to pre-train the reader and retriever components for OpenQA. Similarly,~\cite{lewis2020pre} perform cross-lingual pre-training where the objective is to predict a sequence using its paraphrases in different languages, demonstrating improved zero-shot performance in document translation tasks.


\section{Conclusion} \label{sec:conclusion}
We propose approaches to improve the retrieval accuracy of the dual-encoder model for the OpenQA task.
We first perform a systematic investigation of the importance of pre-training with ICT and masked salient spans tasks for supervised training of the retriever.
We then present two approaches for end-to-end training of the reader and retriever components in OpenQA.
In one approach, the reader considers each retrieved document individually while in the other approach where the reader considers all the retrieved documents jointly.
Overall, these methods help achieve state-of-the-art results on both retrieval and answer extraction.


\section*{Acknowledgements}
This work was done during the first author's internship at NVIDIA. It was also partially supported by Canada CIFAR AI Chair held by Prof.\ Hamilton.
We would like to thank the anonymous reviewers for providing valuable feedback and recommendations.
We would also like to thank the administrators of the \href{https://www.top500.org/system/179842/}{Selene} supercomputer for  their  assistance in facilitating the large-scale runs.


\section*{Broader Impact and Ethics Statement}

To understand the ethical context of our work on open-domain question answering, it is important to consider the real-world use cases and potential individuals who may interact with systems developed based on our proposed methods. The potential real-world applications could be search engines or virtual assistants, where our techniques can improve the question-answering ability. However, it is worthwhile to mention that our trained systems can not be deployed off-the-shelf for such applications, given that our models were trained on the Natural Questions and TriviaQA datasets with the goal of matching the specific training data distribution. Real-world applications building on our work should be re-trained using a custom training dataset that is relevant to the kind of queries that originates in practice.

Our system represents a prototype model for answering questions over Wikipedia and can easily be extended to be used in sensitive contexts such as legal or health-care settings. However, extensive and robust quality assurance testing will be needed as our system was not designed to meet those criteria. More generally, there is the possibility of social biases which could be introduced by the training data. Since we did not control or regularize our model to remove such biases, we would urge the users to undertake the necessary quality-assurance testing to evaluate and understand the extent to which such biases might be present. User should also understand how much these biases are impacting their trained system and to make modifications to their training data and procedures accordingly.

\bibliography{main.bib}
\bibliographystyle{acl_natbib}

\clearpage
\appendix

\section{Training Details} \label{sec:training_details}
We provide the training details of all the experiments below. We use the same training settings for both the base and large model configurations and use the open-source Megatron-LM toolkit~\cite{shoeybi2019megatron} to implement the models.\footnote{\url{https://github.com/NVIDIA/Megatron-LM}} To train the models, we employed mixed-precision training~\cite{micikevicius2018mixed} and leveraged distributed training feature as implemented in the Pytorch framework~\cite{shen2020pytorch}. All of our experiments were performed on the \href{https://blogs.nvidia.com/blog/2020/06/22/top500-isc-supercomputing/}{Selene} cluster which consists of NVIDIA A100 GPUs.

\subsection{Language Models Training}
We train BERT~\cite{devlin2019bert,Lan2020ALBERT} and T5~\cite{raffel2020t5} language models from scratch, whose hyperparameters for both the base and large configurations are detailed in Table~\ref{table:bert-t5}. We used 32 GPUs to train the BERT-large (330M) model and 128 GPUs to train the T5-large (770M) model.

\begin{table*}[t]
\small
\centering
\begin{tabular}{l c c}
 \toprule
 \textbf{Hyperparameter} & \textbf{BERT} & \textbf{T5} \\
 \midrule
 Dataset & Wikipedia, BookCorpus & Wikipedia, CC-Stories, RealNews, OpenWebText \\
 Hidden Size & \{768, 1024\} & \{768, 1024\} \\
 Attention Heads & \{12, 16\} & \{12, 16\} \\
 Dropout & 0.1 & 0.1 \\
 Attention Dropout & 0.1 & 0.1 \\
 Optimizer & Adam & Adam \\
 Training Steps & 1M & 1M \\
 Warmup Steps & 10k & 10k \\
 Peak Learning Rate & 1e-4 & 1e-4 \\
 Weight Decay & 1e-2 & 1e-2 \\
 Batch Size & 256 & 2048 \\
 Learning Rate Decay & Linear & Linear \\
 Gradient Clipping & 1.0 & 1.0 \\
 \bottomrule
\end{tabular}
\caption{Hyperparameters for pre-training BERT and T5 models.}
\label{table:bert-t5}
\end{table*}

\subsection{Retriever Training}

\paragraph{Supervised:}
We use Adam optimizer~\cite{kingma2014adam}, a batch size of $128$, learning rate of 2e-5 with a linear decay, and train for $80$ epochs. Training was performed on 16 GPUs.

\paragraph{ICT training:}
We initialize the parameters of both the question and context encoders using BERT weights trained with Megatron-LM.
We train the model on Wikipedia paragraphs with maximum length of 256 tokens.
We use a batch size of $4,096$, learning rate of 1e-4 with linear decay, and train the model for $100,000$ steps using Adam optimizer. This corresponds to training the model for roughly $20$ epochs over the Wikipedia dataset.
We set the weight decay to $0.01$ and the warmup ratio of the optimizer to $0.01$. With a probability of $0.1$, we also keep the query sentence in the context.
We train the large ICT model using 128 GPUs.

\paragraph{Masked salient spans generative training:}
We initialize the retriever with ICT training and pre-train the T5 reader on an aggregated dataset from~\cite{shoeybi2019megatron}. We use the pre-trained models provided by the Stanza toolkit~\cite{qi2020stanza} to segment Wikipedia paragraphs into sentences and extract named entities.\footnote{We use the model trained on OntoNotes~\cite{pradhan2012conll} to extract named entities for $10$ selected categories.} The masked sentence is used as a query to retrieve evidence documents with the help of which the reader predicts the masked words. The model is trained according to Equation~\ref{eq:indiv-topk} and~\ref{eq:loss}.
We train the model for $100,000$ steps with Adam optimizer using a learning rate of 2e-5 and a warmup ratio of $0.05$. 
Similar to~\cite{guu2020realm}, we also compute the evidence embeddings asynchronously and update the evidence index every $500$ steps. Training was performed on 240 GPUs.

\subsection{End-to-End Supervised Training}
As the performance of the ICT pre-trained retriever and masked salient spans pre-trained retriever is similar when all the training data is used (Sec.~\ref{subsec:ret-init}), we select the retriever pre-trained with ICT initialization and finetuned with supervised data. For the reader, we use a pre-trained T5 model. For all experiments, we train for $10$ epochs using a batch size of $64$, learning rate of 2e-5 with linear decay, and weight regularization of $0.1$. For \emph{Individual Top-k} approach, during training, the evidence embeddings index is refreshed after every $500$ steps. The number of retrieved evidence documents for the reader is considered as a hyperparameter and is selected via performance on the dev set. Training of Individual Top-k was performed on 240 GPUs while training of Joint Top-k was performed on 64 GPUs.

For retrieving the top-k documents from our evidence ($\sim$21M documents), we perform exact search. Specifically, we utilize matrix multiplication and top-k functionalities as provided by the PyTorch framework. This matrix multiplication operation is highly optimized for GPU computations and
we observed that performing exact search was not a bottleneck during training. We therefore did not optimize or approximate the similarity search using LSH~\cite{andoni2015lsh} or efficient maximum inner product search~\cite{srivastava2014lsh}.

\paragraph{NQ and TriviaQA Specific Details:} For both datasets, we uniformly sample the target answer from the list of provided answers during the training process. For answer extraction, similar to ~\cite{guu2020realm}, we did not append the title of the Wikipedia article with the corresponding top-k retrieved document as the reader's input.

\subsection{Individual Top-k Inference}
During inference, the reader model first greedily generates an answer for each retrieved document. We then score each generated answer using Eq.~\ref{eq:indiv-topk} and finally select the answer with the highest likelihood score. 

\subsection{Example Outputs from Retriever}
We present few examples in Table~\ref{table:examples} when the ICT + Supervised retriever is evaluated on the NQ test dataset.

\begin{table*}[t]
\begin{minipage}{\linewidth}
\small
\centering
\begin{tabularx}{\linewidth}{@{} X l X @{}}
\toprule
\textbf{Question from NQ test} & \textbf{Answer} & \textbf{Top-1 Document Retrieved by ICT + Supervised} \\
\midrule
what parts make up the peripheral nervous system & autonomic nervous system & \ldots The connection between CNS and organs allows the system to be in two different functional states: sympathetic and parasympathetic. The peripheral nervous system is divided into the somatic nervous system, and the \textbf{autonomic nervous system}. The somatic nervous system is under voluntary control, and transmits signals from the brain to end organs such as muscles. The sensory nervous system is part of the somatic nervous system and transmits signals from senses such as taste and touch (including fine touch and gross touch) to the spinal cord and brain\ldots \\ 
\midrule
when is the new season of wentworth coming out & 19 June 2018 & \ldots In a similar manner, a 12-episode fourth season was announced before the airing of the third season on 27 February 2015. It began airing from 10 May 2016. Cormack confirmed a fifth season had been commissioned on 19 July. The twelve-part series premiered on 4 April 2017. On 9 May 2017, Showcase announced that the series has been renewed for a sixth season, which premiered on \textbf{19 June 2018}. A seventh season was commissioned in April 2018, before the sixth-season premiere, with filming commencing the following week and a premiere set for 2019\ldots \\
\midrule
who challenged the aristotelian model of a geocentric universe & Copernicus & \ldots ("On the Revolutions of the Heavenly Spheres"), which posited that the Earth and the other planets instead revolved around the Sun. The geocentric system was still held for many years afterwards, as at the time the Copernican system did not offer better predictions than the geocentric system, and it posed problems for both natural philosophy and scripture. The Copernican system was no more accurate than Ptolemy\'s system, because it still used circular orbits. This was not altered until Johannes Kepler postulated that they were elliptical (Kepler\'s first law of planetary motion). \ldots \\
\bottomrule
\end{tabularx}
\caption{Examples of top-1 retrieved documents from the NQ test as outputted from the  ICT + Supervised retriever. If the answer exists in the document, it is highlighted in bold.}
\label{table:examples}
\end{minipage}
\end{table*}

\section{Reproducibility Checklist}

\subsection{For all reported experimental results}
\begin{itemize}
	\item \textit{A clear description of the mathematical setting, algorithm, and/or model}: This is provided in the main paper in Sec.~\ref{sec:retriever} and Sec.~\ref{sec:end-to-end-training}.

	\item \textit{A link to a downloadable source code, with specification of all dependencies, including external libraries (recommended for camera ready, though welcome for initial submission)}: As mentioned previously, we have developed our codebase over the open-source Megatron-LM library (\url{https://github.com/NVIDIA/Megatron-LM}). Our implementations over this codebase are currently organized in different branches, that are better suited for walk-through with a git-based tool. To preserve anonymity and in good faith, we are submitting the source codes from one branch of our codebase, with the caution that the codebase doesn't contain an exhaustive README file.

	\item \textit{A description of computing infrastructure used}: We run experiments on Nvidia's Selene cluster where each node's specifications are: Number of CPUs: 256, Physical Memory: 2.2TB, GPU model: 8 x Nvidia A100, GPU architecture and memory: Ampere/80GB, Arch: x86\_64, and Disk size: 10TB.

	\item \textit{The average runtime for each model or algorithm, or estimated energy cost}: We provide the average runtime and compute used for training different models in Appendix~\ref{sec:training_details}. However, we want to highlight that our codes were not carefully optimized to minimize runtime or to make optimal use of the hardware resources.

	\item \textit{The number of parameters in each model}: We provide number of parameters in models in Sec.~\ref{sub:model_details}.

	\item \textit{Corresponding validation performance for each reported test result}: Validation set performance is currently not reported in the main paper. However, we followed rigorous experimentation protocol, and selected the best models by its performance on the validation set. If the program committee or reviewers require the validation set performance, we will include it in the final version of the paper.
	
	\item \textit{A clear definition of the specific evaluation measure or statistics used to report results}: Our evaluation metrics are standard and widely used by the question answering community. We provide their details in the main paper in Sec.~\ref{results-ret-training} and Sec.~\ref{sec:answer-extraction}.
\end{itemize}

\subsection{For all results involving multiple experiments, such as hyperparameter search}
\begin{itemize}
	\item \textit{The exact number of training and evaluation runs:} We provide training details for all models in Appendix~\ref{sec:training_details}. Specifically, for the finetuning experiments, we train the models until convergence, which is $80$ epochs for retriever models and $10$ epochs for answer extraction models. We evaluate the model after each epoch on the validation set and save the best checkpoint according to their performance on the corresponding evaluation metric.

	\item \textit{Hyperparameter configurations for best-performing models}: We provide the hyperparameter settings in Appendix~\ref{sec:training_details}.
	
	\item \textit{The bounds for each hyperparameter}: As described in Appendix~\ref{sec:training_details}, our model and training setting uses standard hyperparameters such as different dropouts $\in [0, 1)$, warmup ratio of optimizer $\in [0.01, 0.05]$, weight regularization $\in [0, 1]$, and learning rate $\in [1e^{-4}, 1e^{-5}]$. The model hyperparameters includes model dimensions $d \in \{768, 1024\}$, number of layers $\in \{12, 24\}$.
	
	\item \textit{The method of choosing hyperparameter values (e.g., uniform sampling, manual tuning, etc.) and the criterion used to select among them (e.g., accuracy)}: We performed manual hyperparameter tuning. We also performed tuning of the number of warmup steps for the Adam optimizer. We selected the best hyperparameter using performance on the validation set.
	
	\item \textit{Summary statistics of the results (e.g. mean, variance, error bars, etc.)}: All of our experiments are compute expensive large-scale runs utilizing a lot of resources such as CPUs, GPUs and take time ranging from tens of hours to several days. Therefore, due to computational and time constraints performing multiple runs for each experiment was not feasible. Therefore, we adopted the approach of using the same seed value (1234) for all the training runs including both pre-training and finetuning experiments.
\end{itemize}

\subsection{For all datasets used}
\begin{itemize}
    \item \textit{Details of train/validation/test splits}: We use the standard training / dev / test splits whose details are provided in Sec.~\ref{sec:exp-setup}.

	\item \textit{Relevant statistics such as number of examples and label distributions}: We provide dataset statistics details in Table~\ref{tab:dataset_stats}.

	\item \textit{An explanation of any data that were excluded, and all pre-processing steps}: We include the relevant details in Sec.~\ref{sec:exp-setup}.

	\item \textit{For natural language data, the name of the language(s)}: Our datasets are in English language.
	
	\item \textit{A link to a downloadable version of the dataset or simulation environment}: Both the datasets of NQ and TriviaQA are open-source and widely used by the community. NQ is available at: \url{https://ai.google.com/research/NaturalQuestions/download}. TriviaQA is available at: \url{http://nlp.cs.washington.edu/triviaqa/}. We make use of the NQ, TriviaQA, and Wikipedia datasets as open-sourced by the DPR authors~\cite{karpukhin2020dense} here: \url{https://github.com/facebookresearch/DPR/blob/master/data/download_data.py}.
	
	\item \textit{For new data collected, a complete description of the data collection process, such as instructions to annotators and methods for quality control}: This is not applicable to this work.
\end{itemize}

\end{document}